\documentclass{article} 
\usepackage[preprint]{colm2026_conference}
\usepackage{microtype}
\usepackage{hyperref}
\usepackage{url}
\usepackage{booktabs}
\usepackage{adjustbox}
\usepackage{multirow}
\usepackage{newfloat}
\usepackage{listings}
\usepackage{enumitem}
\usepackage{amsmath}
\usepackage{amssymb}
\usepackage{amsthm}
\usepackage{caption}
\usepackage{colortbl}
\usepackage{xcolor}
\usepackage{color}
\usepackage{tabularx}  
\usepackage{array}     
\usepackage{ulem}

\usepackage{graphicx}
\usepackage{xspace}
\usepackage{algorithm}
\usepackage{algpseudocode}
\usepackage{longtable}
\usepackage{arydshln}
\usepackage{wrapfig}
\usepackage[listings]{tcolorbox}

\usepackage{lineno}
\usepackage[dvipsnames]{xcolor}

\definecolor{darkblue}{rgb}{0, 0, 0.5}
\hypersetup{colorlinks=true, citecolor=darkblue, linkcolor=darkblue, urlcolor=darkblue}

\title{Improving Search Agent with One Line of Code}

\author{
 \textbf{Jian Li\textsuperscript{1,2}\textsuperscript{*}\textsuperscript{$\dag$}},
  \textbf{Dongsheng Chen\textsuperscript{2}},
  \textbf{Zhenhua Xu\textsuperscript{2}},
 \textbf{Yizhang Jin\textsuperscript{2}},
 \textbf{Jiafu Wu\textsuperscript{2}},
 \\
 \textbf{Chengjie Wang\textsuperscript{2}},
 \textbf{Xiaotong Yuan\textsuperscript{1}\textsuperscript{$\dag$}},
 \textbf{Yabiao Wang\textsuperscript{2}\textsuperscript{*}}
\\
 \textsuperscript{1}Nanjing University,
 \textsuperscript{2}Tencent YoutuLab,
\\
 \small{
   $^*$ Project leader. $^\dagger$ Corresponding author.
 }
}

\begin{document}

\ifcolmsubmission
\linenumbers
\fi

\maketitle

\begin{abstract}

Tool-based Agentic Reinforcement Learning (TARL) has emerged as a promising paradigm for training search agents to interact with external tools for multi-turn information-seeking process autonomously .
However, we identify a critical training instability that leads to catastrophic model collapse: Importance Sampling Distribution Drift (ISDD).
In Group Relative Policy Optimization (GRPO), a widely adopted TARL algorithm, ISDD manifests as a precipitous decline in the importance sampling ratios, which nullifies gradient updates and triggers irreversible training failure.
To address this, we propose \textbf{S}earch \textbf{A}gent \textbf{P}olicy \textbf{O}ptimization (\textbf{SAPO}), which stabilizes training via a conditional token-level KL constraint. 
Unlike hard clipping, which ignores distributional divergence, SAPO selectively penalizes the KL divergence between the current and old policies. Crucially, this penalty is applied only to positive tokens with low probabilities where the policy has shifted excessively, thereby preventing distribution drift while preserving gradient flow.
Remarkably, SAPO requires only one-line code modification to standard GRPO, ensuring immediate deployability. 
Extensive experiments across seven QA benchmarks demonstrate that SAPO achieves \textbf{+10.6\% absolute improvement} (+31.5\% relative) over Search-R1, yielding consistent gains across varying model scales (1.5B–14B) and families (Qwen, LLaMA).
\end{abstract}

\section{Introduction}

Recent studies~\cite{li2025aissurvey,guo2024agentsurvey} have introduced search agents capable of interpreting user intent, planning strategies, executing multi-turn actions, and accumulating retrieved evidence for answer generation. Representative methods, such as Search-R1~\cite{jin2025searchr1}, employ Group Relative Policy Optimization (GRPO) ~\cite{kaelbling1996rl,shao2024deepseekmath} with rule-based rewards to train agents to interact with a search tool over large corpora. Consequently, GRPO has emerged as a prevalent post-training algorithm for Tool-based Agentic Reinforcement Learning (TARL), inspiring numerous follow-up works ~\cite{mei20252o2searcher,shi2025autorefine,qian2025inforage,zhang2025criticsearch}.

Despite promising results, existing TARL frameworks face a critical limitation: \textbf{standard GRPO introduces Importance Sampling Distribution Drift (ISDD) when the policy shifts outside the trust region.} While several group-based optimization methods ~\citep{DAPO, SimpleRL-Zoo, Skywork-OR1, CISPO, POLARIS, Archer,zhang2025survey} have adopted PPO\_clip~\cite{schulman2017ppo} to restrict the deviation between current and old policies. ISDD remains a persistent phenomenon in search agent training, particularly as response lengths or search turns increase. In GRPO, all tokens within a response share an identical advantage value. 
leading to two primary issues: (1) \textit{Numerical Inaccuracy}: The advantage value may not reflect the true contribution of individual tokens, as different steps contribute unequally to the final answer; and (2) \textit{Misleading Updates}: A response with a correct final answer may contain incorrect intermediate steps (and vice versa). 
Conflicts arise when the old policy assigns high probability to intermediate correct tokens, but the current policy suppresses them due to negative advantage, leading to vanishing importance sampling weights.

\begin{figure*}[h]
	\centering
	\includegraphics[width=0.98\linewidth]{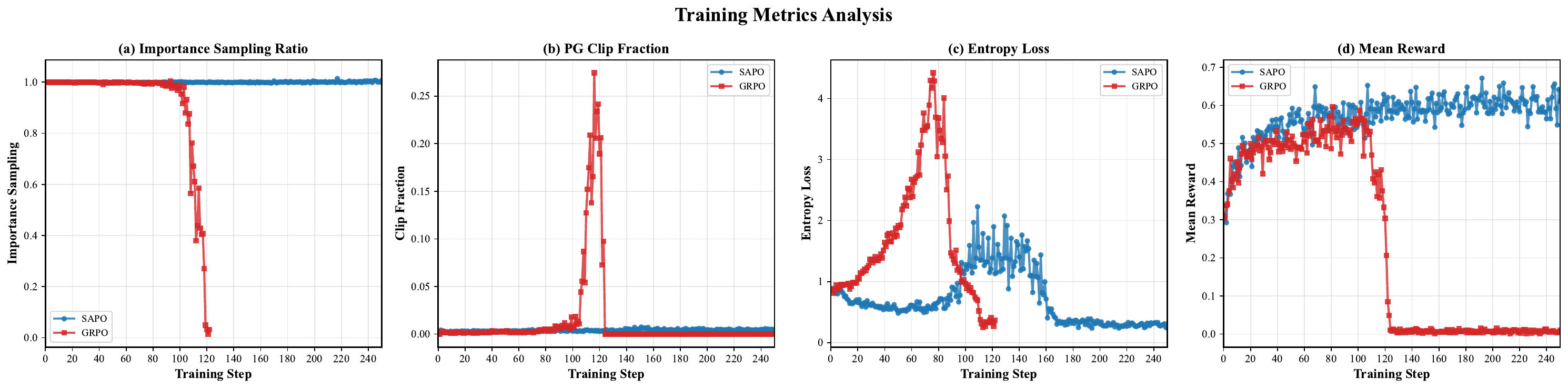}
	\caption{Comparison of training dynamics between \textbf{SAPO} and GRPO regarding (a) Importance Sampling Ratio, (b) Clip Ratio, (c) Entropy, and (d) Reward.}
	\label{fig1}
\end{figure*}

Figure.\ref{fig1} depicts the training dynamics of Search-R1 using GRPO, tracking the Importance Sampling (IS) ratios, clip ratio, entropy, and reward. While the IS ratios remain near 1 during the initial phase, they subsequently drop sharply, indicating a growing divergence between the current and old policies. Consequently, the clip ratio exhibits a pronounced spike in the latter stages. Similarly, policy entropy rises substantially at the beginning, suggesting the presence of low-probability positive tokens or high-probability negative tokens. Given the convergence of the IS ratios toward zero, we identify \textbf{the suppression of positive tokens (low-probability positive actions)} as the primary driver of ISDD. While the outcome-based reward initially improves, it deteriorates as the IS ratios destabilize and entropy collapses, indicating that late-stage instability compromises final performance. These empirical results demonstrate that despite the use of hard clipping, catastrophic and irreversible model collapse persists.

The primary motivation behind IS and clipping is to correct for distribution shifts and ensure stability. However, for the positive tokens, if the probability under the current policy drops significantly below that of the old policy, the update is effectively suppressed by excessively low importance weights. This yields a weak update signal, preventing the model from recovering valid actions. This observation prompts the following research question:
\begin{center}
\textit{How can we mitigate optimization conflicts using appropriate constraints \\ when the current policy diverges significantly from the old policy on positive tokens.}
\end{center}

To enhance training stability, we propose \textbf{S}earch \textbf{A}gent \textbf{P}olicy \textbf{O}ptimization (SAPO), a simple yet theoretically grounded modification of GRPO. Following PPO\_KL ~\cite{schulman2017ppo}, SAPO introduces a conditional KL penalty term to enforce a token-level constraint on the distributional divergence between the current and old policies. To optimize constraint efficiency, SAPO employs an asymmetric mechanism that selectively penalizes conflicting positive tokens and preserves informative gradients from positive tokens. Consequently, SAPO functions as a soft trust region constraint: instead of hard-clipping, we softly penalize large IS ratios, allowing adaptive step sizes while maintaining distributional proximity.

We summarize our contributions as follows. 

\begin{itemize}

\item We propose SAPO, a policy optimization method designed to stabilize post-training for autonomous multi-turn search agents tackling complex, real-world questions. 

\item We introduce a conditional KL penalty term that enforces a token-level constraint on the distributional divergence, specifically targeting low-probability positive tokens.

\item We demonstrate the effectiveness and generalizability of SAPO across diverse search agents, achieving improvement performance on seven challenging QA benchmarks.
\end{itemize}

\section{Preliminaries}

\subsection{Proximal Policy Optimization}
\label{sec:ppo}
Proximal Policy Optimization (PPO) ~\cite{schulman2017ppo} is one of the most prominent policy gradient methods in reinforcement learning (RL)~\cite{kaelbling1996rl}. PPO facilitates stable training by utilizing IS for off-policy updates while constraining the deviation between the current and old policies. The standard unclipped PPO objective is defined as:

\begin{equation}
\begin{aligned}
\mathcal{J}_{\text{PG}}(\theta) &= 
\mathbb{E}_{q \sim \mathcal{D},\, o \sim \pi_{\theta_{\text{old}}}(\cdot \mid q)} 
\left[ r_t(\theta) A_t \right], & \text{where} \quad
r_t(\theta) = \frac{\pi_\theta(o_t \mid q, o_{<t})}
{\pi_{\theta_\text{old}}(o_t \mid q, o_{<t})}
\end{aligned}
\end{equation}

Here, $\pi_{\theta_{\text{old}}}$ and $\pi_{\theta}$ denote the old policy and the current policy, respectively, while $A_t$ represents the advantage function. $q$ denotes a query sampled from the data distribution $\mathcal{D}$, and $o$ represents a response generated by the old policy. Direct optimization of this objective can lead to excessive deviation in the IS ratios $r_t(\theta)$, resulting in training instability. To mitigate this, PPO introduces a clipped surrogate objective that imposes a local trust region:

\begin{equation}
\begin{aligned}
\mathcal{J}_{\text{PPO}}(\theta) =
\mathbb{E}_{q \sim \mathcal{D},\, o \sim \pi_{\theta_{\text{old}}}(\cdot \mid q)}  \left[
\min \left(
r_t(\theta) A_t,\,
\text{clip}\!\left(r_t(\theta), 1 - \epsilon, 1 + \epsilon \right) A_t
\right)
\right]
\end{aligned}
\end{equation}

Here, $\epsilon$ is a hyperparameter defining the clipping bounds. This mechanism limits the incentive for the policy to update when the IS ratio deviates significantly, thereby enhancing stability. However, it zeros out the gradient for clipped tokens, potentially allowing the policy to drift beyond the intended trust region without correction.





\subsection{Group Relative Policy Optimization}
\label{sec:grpo}

Group Relative Policy Optimization (GRPO)~\cite{guo2025deepseekr1} eliminates the need for a value function critic by employing group-based advantage estimation. Following Search-R1 ~\cite{jin2025searchr1}, External retrieval tokens are masked during search agent training. Consequently, the GRPO objective is optimized exclusively over the agent-generated tokens:

\begin{small}
\begin{equation}
    \label{eq:grpo}
    \begin{aligned}
    \mathcal{J}_{\text{GRPO}}(\theta) =  &\mathbb{E}_{q \sim \mathcal{D}, \{o_i\}_{i=1}^G \sim \pi_{\theta_{\text{old}}}(\cdot|q)}
    \left[
    \frac{1}{G} \sum_{i=1}^{G} 
    \frac{1}{|o_i|} \sum_{t=1}^{|o_i|}
    \min \Big(
     r_{i,t}(\theta)
     \hat{A}_{i,t},
    \text{clip} \left(
    r_{i,t}(\theta)
    , 1-\epsilon, 1+\epsilon
    \right) \hat{A}_{i,t}
    \Big) 
    \right]
    \end{aligned}
\end{equation}
\end{small}

where the IS ratio $r_{i,t}(\theta)$ and the advantage $\hat{A}_{i,t}$ are defined as:

\begin{equation}
\label{eq:IS ratio}
r_{i,t}(\theta) = \frac{ \pi_{\theta}(o_{i,t} \mid q, o_{i,<t}) }
     { \pi_{\theta_{\text{old}}}(o_{i,t} \mid q, o_{i,<t}) }, \qquad
\hat{A}_{i,t} = \hat{A}_{i} = \frac{ R_{i} - \text{mean}(\{R_j\}_{j=1}^G) }  {\text{std}(\{R_j\}_{j=1}^G))} 
\end{equation}

Here, the advantage is computed by normalizing the sequence-level reward $R_{i}$ against the group statistics. Notably, GRPO assigns the same advantage value to all tokens.

\section{Approach}

\subsection{Search Agent Formulation}

Let $q$ denote the user question, $a$ the generated answer, and $\pi_{\theta}$ the Large Language Model(LLM). We formulate the standard question-answering task as $a = \pi_{\theta}(\cdot \mid q)$.
 
RAG systems enhance this process by retrieving knowledge $k$ from an external corpus $\mathcal{D}$ using a retriever $\mathcal{R}(\cdot)$, subsequently generating an answer conditioned on both $q$ and $k$. We formulate RAG as $a = \pi_{\theta}(q \mid k), \quad k = \mathcal{R}(q , \mathcal{D})$.

However, this static workflow is often insufficient for complex search tasks that necessitate iterative reasoning and adaptive retrieval. Recent advancements in search agents enable a more dynamic paradigm where retrieval is interleaved with reasoning.

\begin{equation}
a = \pi_{\theta}\left(q \mid  \{z_i , k_i\}_{i=1}^{T}\right),  
k_i = \mathcal{R}(sq_i \mid \mathcal{D}),  sq_i \in z_i
\end{equation}

Here, the policy output $o$ comprises intermediate reasoning steps $z$ and the final answer $a$. $sq_i$ denotes a search query generated within the reasoning step $z_i$ to retrieve intermediate knowledge $k_i$. The sequence $\{z_i, k_i\}$ represents the interleaved trajectory of reasoning and retrieval knowledge. $T$ is the number of search actions.

\subsection{Importance Sampling Distribution Drift}

In search agents, optimal strategies are inherently sparse, as only a limited subset of queries yields effective results. Consequently, When the policy $\pi_\theta$ rapidly diverges from the exploration path of $\pi_{\theta_{\text{old}}}$, it may assign negligible probabilities to the sampled actions, causing $\pi_\theta(o_t)$ to be orders of magnitude smaller than $\pi_{\theta_{\text{old}}}(o_t)$, driving $r_t(\theta) \to 0$.

\theoremstyle{definition} 
\newtheorem{definition}{Definition} 

\begin{definition}[ISDD]
\label{def:isdd}
Building upon the IS ratios defined in Eq.~\ref{eq:IS ratio}, we define Importance Sampling Distribution Drift (ISDD) as the event where:
\begin{equation}
\mathbb{P}\left( \prod_{t=1}^{T} r_t(\theta) < \epsilon \right) > \phi
\end{equation}
for small $\epsilon > 0$ and a probability threshold $\phi$. This condition indicates that a significant proportion of trajectories suffer from vanishing cumulative importance weights.
\end{definition}

\paragraph{Why ISDD Causes Collapse.}
In GRPO, the policy gradient is weighted by IS ratios:

\begin{equation}
    \label{eq:grpo3}
    \begin{aligned}
    \nabla_{\theta} \mathcal{J}_\text{GRPO} (\theta) 
    =&\ 
    \mathbb{E}_{q \sim \mathcal{D}, \{o_i\}_{i=1}^G \sim \pi_{\theta_{\text{old}}}(\cdot|q)}
        \Bigg[
        \frac{1}{G} \sum_{i=1}^{G} 
        \frac{1}{|o_i|} \sum_{t=1}^{|o_i|} \hat{A}_{i,t}r_{i,t}(\theta) \nabla_{\theta} \log \pi_{\theta} (o_{i,t} | q, o_{i,<t}) 
    \Bigg]
    \end{aligned}
\end{equation}

When ISDD occurs ($r_t \to 0$), the gradients vanish regardless of the advantage values, freezing the policy update. Consequently, even high-reward trajectories contribute negligible gradients, hindering the model's ability to learn from successful explorations.

\theoremstyle{plain} 
\newtheorem{proposition}{Proposition}

\begin{proposition}[ISDD Amplification in Interleaved Multi-Step Interactions]
\label{prop:isdd-interleaved}

Let the IS ratios $r_t$ follow a log-normal distribution $\log r_t \sim \mathcal{N}(\mu, \sigma^2)$, with the drift parameter $\lambda = \mu + {\sigma^2}/{2}$.
Due to the low-entropy and bottleneck nature of tool selection, action tokens $a_i$ exhibit significantly higher sensitivity to policy shifts than reasoning tokens $z_i$, implying $\lambda_{k} \ll \lambda_a < 0$.

The expected cumulative importance weight for the trajectory decomposes as follows:
\begin{equation}
\label{eq:isdd_interleaved}
\mathbb{E}\left[\prod_{t=1}^{L} r_t\right] = \mathbb{E}\left[\prod_{t \in \mathcal{Z}} r_t\right] \cdot \mathbb{E}\left[\prod_{t \in \mathcal{A}} r_t\right] = \underbrace{\exp\left(L_z \lambda_z\right)}_{\text{Reasoning Drift}} \cdot \underbrace{\exp\left(L_a \lambda_a\right)}_{\text{Interaction Drift}}
\end{equation}
where $L$ denote the total number of tokens.
ISDD is more severe in agent tasks than in QA tasks because the \textbf{Interaction Drift} term decays exponentially with the volume of actions $L_a$, driven by the highly negative drift parameter $\lambda_a$.
\end{proposition}


\subsection{Search Agent Policy Optimization}
To alleviate ISDD and enhance training stability, SAPO introduces an auxiliary penalty term. This term enforces a constraint on the distributional divergence between the current and old policies. The SAPO token-level objective is formulated as:

\begin{equation}
    \label{eq:sapo_kl_def}
    \begin{aligned}
    \mathcal{J}_{\text{SAPO}}(\theta) =  &\mathbb{E}_{q \sim \mathcal{D}, \{o_i\}_{i=1}^G \sim \pi_{\theta_{\text{old}}}(\cdot|q)}
    \Bigg[
    \frac{1}{G} \sum_{i=1}^{G} 
    \frac{1}{|o_i|} \sum_{t=1}^{|o_i|} \\&
    \min \Big(
     r_{i,t}(\theta)
     \hat{A}_{i,t},
    \text{clip} \left(
    r_{i,t}(\theta)
    , 1-\epsilon, 1+\epsilon
    \right) \hat{A}_{i,t}
    \Big) 
   \underbrace{+ \gamma \text{KL} \left[ \pi_{\theta} \, \| \, \pi_{\text{old}}\right] }_{\textcolor{BrickRed}{\text{Drift Penalty}}}
    \Bigg]
    \end{aligned}
\end{equation}

where $\gamma$ denotes the penalty coefficient. Unlike the KL penalty in standard RL, which typically regulates divergence against a fixed reference model, SAPO penalizes deviations from the dynamic old policy to prevent the vanishing gradient problem. We approximate the KL divergence using the log-ratio:

\begin{equation}
    \label{eq:grpo}
    \begin{aligned}
    \text{KL} \left[ \pi_{\theta} \, \| \, \pi_{\text{old}} \right] = \log \frac{ \pi_{\theta}(o_{i,t} \mid q, o_{i,<t}) }{ \pi_{\theta_{\text{old}}}(o_{i,t} \mid q, o_{i,<t}) } = \log r_{i,t}(\theta)
    \end{aligned}
\end{equation}

Standard KL penalties may indiscriminately hinder policy exploration. To address this, we propose a conditional KL penalty that activates only when the probability ratio drops significantly for samples with positive advantages. The conditional term is defined as:

\begin{equation}
    \label{eq:grpo}
    \begin{aligned}
    \text{KL}_{cond} \left[ \pi_{\theta} \, \| \, \pi_{\text{old}} \right] =  \mathbb{I}(r_{i,t}(\theta)<\tau, \hat{A}_{i,t}>0 ) \log r_{i,t}(\theta)
    \end{aligned}
\end{equation}

Here, $\mathbb{I}(\cdot)$ is the indicator function and $\tau$ is the threshold to identify drifting tokens. The penalty term has three crucial properties:
\begin{enumerate}
\item \textbf{Conditional}: Only applied when $A_t > 0$ (positive advantage)
\item \textbf{Threshold-gated}: Only triggered when $r_t < \tau$ (excessive shift)
\item \textbf{Logarithmic}: $\log r_t$ grows slowly, allowing gradual exploration
\end{enumerate}


\subsection{Gradient Analysis}
\label{subsec:gradient}

We derive the gradient of the SAPO auxiliary term as follows:

\begin{small}
\begin{equation}
    \label{eq:sapo_kl_grad}
    \begin{aligned}
     \nabla_{\theta} \mathcal{J}_\text{SAPO\_KL} (\theta) 
     = &\ 
     \nabla_{\theta}\mathbb{I}(r_{i,t}(\theta)<\tau, \hat{A}_{i,t}>0 )\log r_{i,t}(\theta)  \\
     =&\ 
      \mathbb{I}(\hat{A}_{i,t}>0) \nabla_{\theta} \left[\mathbb{I}(r_{i,t}(\theta)<\tau)\log r_{i,t}(\theta) \right]  \\
     =&\ 
      \mathbb{I}(\hat{A}_{i,t}>0)[ \mathbb{I}(r_{i,t}(\theta)<\tau) \nabla_{\theta}\log r_{i,t}(\theta) + \nabla_{\theta} \mathbb{I}(r_{i,t}(\theta)<\tau) \log r_{i,t}(\theta)]  \\
     =&\ 
      [\mathbb{I}(r_{i,t}(\theta)<\tau, \hat{A}_{i,t}>0) - \delta(r_{i,t}(\theta) - \tau) r_{i,t}(\theta) \log r_{i,t}(\theta) ]  \nabla_{\theta}\log \pi_{\theta} (o_{i,t} | q, o_{i,<t})  \\
    \end{aligned}
\end{equation}
\end{small}

Consequently, the total gradient of the SAPO objective is:

\begin{small}
\begin{equation}
    \label{eq:grpo3}
    \begin{aligned}
    \nabla_{\theta} \mathcal{J}_\text{SAPO} (\theta) 
    =&\ 
    \mathbb{E}_{q \sim \mathcal{D}, \{o_i\}_{i=1}^G \sim \pi_{\theta_{\text{old}}}(\cdot|q)}
        \Bigg[
        \frac{1}{G} \sum_{i=1}^{G} 
        \frac{1}{|o_i|} \sum_{t=1}^{|o_i|} (\hat{A}_{i,t}r_{i,t}(\theta) \\
           &\
         \underbrace{ + \gamma [\mathbb{I}(r_{i,t}(\theta)<\tau, \hat{A}_{i,t}>0) - \delta(r_{i,t}(\theta)-\tau) r_{i,t}(\theta) \log r_{i,t}(\theta) ]}_{\textcolor{BrickRed}{\text{Drift Penalty Gradient Coefficient}}}) \nabla_{\theta} \log \pi_{\theta} (o_{i,t} | q, o_{i,<t})  
    \Bigg]. 
    \end{aligned}
\end{equation}
\end{small}

where $\delta(\cdot)$ denotes the Dirac delta function, arising from the differentiation of $\mathbb{I}(\cdot)$.

\begin{table*}[t]
    \centering
    \small  
    \setlength{\tabcolsep}{4.5pt} 
    \renewcommand{\arraystretch}{1.15} 
    
    \caption{
        Accuracy comparison of SAPO-3B against baseline methods using Qwen2.5-3B \cite{qwen2025qwen25technicalreport} across multiple QA benchmarks. \textbf{Bold} indicates best results, and \uline{underline} denotes second best results.
    }
    \label{tab_exp_main}
    \resizebox{\linewidth}{!}{ 
    \begin{tabular}{lcccccccccc}
        \toprule
        & \multicolumn{4}{c}{\textbf{Single-Hop QA }} & \multicolumn{5}{c}{\textbf{Multi-Hop QA }} & \multirow{2}{*}{\textbf{Avg.}} \\
        \cmidrule(lr){2-5} \cmidrule(lr){6-10} 
        \textbf{Methods} & NQ & TriviaQA & PopQA & \textit{Avg.} & HotpotQA & 2Wiki & Musique & Bamboogle & \textit{Avg.} &  \\
        \midrule
        
        \multicolumn{11}{l}{\textit{\textbf{w/o Retrieval}}} \\
        \quad Direct Generation & 0.106 & 0.288 & 0.108 & 0.167 & 0.149 & 0.244 & 0.020 & 0.024 & 0.109 & 0.134 \\
        \quad SFT & 0.249 & 0.292 & 0.104 & 0.215 & 0.186 & 0.248 & 0.044 & 0.112 & 0.148 & 0.176 \\
        \quad R1-Instruct \cite{guo2025deepseekr1} & 0.210 & 0.449 & 0.171 & 0.277 & 0.208 & 0.275 & 0.060 & 0.192 & 0.184 & 0.224 \\
        \quad R1-Base \cite{guo2025deepseekr1} & 0.226 & 0.455 & 0.173 & 0.285 & 0.201 & 0.268 & 0.055 & 0.224 & 0.187 & 0.229 \\
        \addlinespace 
        
        \multicolumn{11}{l}{\textit{\textbf{Workflow w/ Retrieval}}} \\
        \quad Naive RAG \cite{lewis2020rag} & 0.348 & 0.544 & 0.387 & 0.426 & 0.255 & 0.226 & 0.047 & 0.080 & 0.152 & 0.270 \\
        \quad IRCoT \cite{trivedi2022ircot} & 0.111 & 0.312 & 0.200 & 0.208 & 0.164 & 0.171 & 0.067 & 0.240 & 0.161 & 0.181 \\
        \addlinespace
        
        \multicolumn{11}{l}{\textit{\textbf{Agent w/ Retrieval}}} \\
        \quad Search-o1 \cite{li2025search-o1} & 0.238 & 0.472 & 0.262 & 0.324 & 0.221 & 0.218 & 0.054 & 0.320 & 0.203 & 0.255 \\
        \quad Search-R1-Instruct \cite{jin2025searchr1} & 0.397 & 0.565 & 0.391 & 0.451 & 0.331 & 0.310 & 0.124 & 0.232 & 0.249 & 0.336 \\
        \quad Search-R1-Base \cite{jin2025searchr1} & 0.421 & 0.583 & 0.413 & 0.472 & 0.297 & 0.274 & 0.066 & 0.128 & 0.191 & 0.312 \\
        \quad ReSearch-Instruct \cite{chen2025ReSearch} & 0.365 & 0.571 & 0.395 & 0.444 & 0.351 & 0.272 & 0.095 & 0.266 & 0.246 & 0.331 \\
        \quad ReSearch-Base \cite{chen2025ReSearch} & 0.427 & 0.597 & 0.430 & 0.485 & 0.305 & 0.272 & 0.074 & 0.128 & 0.195 & 0.319 \\
        \quad ZeroSearch-Base \cite{sun2025zerosearch} & 0.430 & 0.616 & 0.414 & 0.487 & 0.338 & 0.346 & 0.130 & 0.139 & 0.238 & 0.345 \\
        \quad StepSearch-Base \cite{wang2025stepsearch} & -- & -- & -- & -- & 0.329 & 0.339 & 0.181 & 0.328 & 0.294 & -- \\
        \quad EXSEARCH-Base \cite{shi2025exsearch} & 0.368 & -- & -- & -- & 0.422 & 0.372 & 0.138 & -- & -- & -- \\
        \quad $O^2$-Searcher \cite{mei20252o2searcher} & 0.444 & 0.597 & 0.429 & 0.490 & 0.388 & 0.374 & 0.160 & 0.344 & 0.317 & 0.391 \\
        \quad AutoRefine-Instruct \cite{shi2025autorefine} & 0.436 & 0.597 & 0.447 & 0.493 & 0.404 & 0.380 & 0.169 & 0.336 & 0.322 & 0.396 \\
        \quad AutoRefine-Base \cite{shi2025autorefine} & 0.467 & 0.620 & 0.450 & 0.512 & 0.405 & 0.393 & 0.157 & 0.344 & 0.325 & 0.405 \\
        \quad InForage \cite{qian2025inforage} & 0.421 & 0.597 & \uline{0.452} & 0.490 & 0.409 & \uline{0.428} & 0.172 & 0.360 & 0.342 & 0.405 \\
        \quad CriticSearch \cite{zhang2025criticsearch} & -- & -- & -- & -- & 0.414 & 0.409 & 0.180 & 0.368 & 0.343 & -- \\
        \quad SE-Search \cite{li2026sesearch} & \textbf{0.475} & 0.624 & 0.423 & 0.507 & 0.450 & 0.361 & 0.183 & 0.424 & 0.355 & 0.420 \\
        \midrule 
        
        \rowcolor{gray!10}
        \quad \textbf{SAPO-3B-Base (Ours)} & \uline{0.474} & \textbf{0.630} & 0.449 & \uline{0.517} & \uline{0.449} & 0.412 & \uline{0.196} & \uline{0.424} & \uline{0.370} & \uline{0.433} \\
        \rowcolor{gray!10}
        \quad \textbf{SAPO-3B-Instruct (Ours)} & 0.469 & \uline{0.629} & \textbf{0.457} & \textbf{0.518} & \textbf{0.456} & \textbf{0.448} & \textbf{0.203} & \textbf{0.432} & \textbf{0.396} & \textbf{0.442} \\
        \bottomrule
    \end{tabular}
    }
   
\end{table*}

\section{Experiments}
\label{sec_exp}

\subsection{Experiment Settings}
\label{sec:exp_setup}

\paragraph{Datasets.}
We evaluate our method on seven diverse QA benchmarks, categorized into single-hop or multi-hop retrieval tasks. The single-hop datasets include Natural Questions (NQ) \cite{kwiatkowski2019NQ}, TriviaQA \cite{joshi2017triviaqa}, and PopQA \cite{mallen2022popqa}. The multi-hop datasets comprise HotpotQA \cite{yang2018hotpotqa}, 2WikiMultihopQA \cite{ho2020wikimultihopqa}, Musique \cite{trivedi2022musique}, and Bamboogle \cite{press2022bamboogle}.
We employ Exact Match (EM) accuracy as the evaluation metric across all datasets. Following Search-R1 \cite{jin2025searchr1}, SAPO is trained on acomposite dataset consisting of NQ and HotpotQA.

\paragraph{Baselines.}
We compare SAPO against three categories of methods:
(1) \textbf{Retrieval-free methods}: direct LLM generation, Supervised Fine-Tuning (SFT), and R1-style training \cite{guo2025deepseekr1} without retrieval;
(2) \textbf{Retrieval-enhanced workflows}: Naive RAG, which retrieves documents based solely on the input question, and IRCoT \cite{trivedi2022ircot}, which interleaves retrieval with Chain-of-Thought;
(3) \textbf{Retrieval-augmented agents}: recent agentic methods including Search-o1 \cite{li2025search-o1}, Search-R1 \cite{jin2025searchr1}, ReSearch \cite{chen2025ReSearch}, ZeroSearch \cite{sun2025zerosearch}, StepSearch \cite{wang2025stepsearch}, EXSEARCH\cite{shi2025exsearch}, $O^2$-Searcher \cite{mei20252o2searcher}, InForage \cite{qian2025inforage}, CriticSearch \cite{zhang2025criticsearch}, AutoRefine \cite{shi2025autorefine} and SE-Search \cite{li2026sesearch}.

\paragraph{Implementation Details.}
We utilize the external corpus \cite{karpukhin2020wikipediadump} used by Search-R1, with E5-base-v2 \cite{wang2022e5} serving as the retrieval engine. Unless otherwise specified, the retriever returns the top-3 documents per query and the maximum number of search turns is set to $T_{max}=5$. We employ the Qwen2.5 series~\cite{qwen2025qwen25technicalreport} as the backbone language model. In the RL phase, we employ a simple rule-based reward function. Specifically, this is an outcome-oriented reward calculated using the F1-score between the predicted and ground-truth answers. the KL penalty coefficient is set to $\gamma=0.1$. The IS ratios threshold is set to $\tau=1.0$.

\subsection{Main Performance}
\label{sec:exp_main}

Table \ref{tab_exp_main} presents the main experimental results comparing SAPO against baseline methods. The {Avg.} column reports the average accuracy. All methods utilize the same retriever, knowledge corpus, training data, and LLM backbone (Qwen2.5-3B).

\textbf{SAPO outperforms other methods.} SAPO surpasses Search-R1 and state-of-the-art methods (\textit{i.e.}, AutoRefine and CriticSearch) across all seven benchmarks. Our method achieves an average EM accuracy of $0.442$, demonstrating that the proposed SAPO post-training strategy substantially enhances the search agent's capabilities. Specifically, SAPO yields a significant absolute improvement of $10.6$ points (a relative gain of $31.5\%$) over the Search-R1 baseline.

\textbf{SAPO demonstrates particularly strong gains on multi-hop QA benchmarks.}
Performance improvements are most pronounced on complex, multi-hop QA tasks. For instance, compared to the CriticSearch, SAPO improves performance on HotpotQA by $4.2$ percentage points (a $10.1\%$ relative improvement) and on Bamboogle by $6.4$ percentage points (a $17.4\%$ relative improvement). Furthermore, across all multi-hop benchmarks, SAPO achieves an average increase of $5.3$ percentage points (a $15.45\%$ relative improvement). compared to Search-R1, SAPO achieves a substantial increase of $14.7$ percentage points (a $24.9\%$ relative improvement) across the four multi-hop QA benchmarks. These gains are attributed to the training stability provided by our proposed auxiliary KL penalty term.

\subsection{Ablation Studies}
We conduct ablation study to evaluate the impact of SAPO's key components. To ensure a fair comparison, all models are trained for the same 300 steps, sampling 5 rollout responses. The baseline, Search-R1, utilizes Qwen2.5-3B-Instruct as the language model and the F1 score as the outcome-based reward. Table \ref{tab_exp_ablation} details the accuracy across seven benchmarks and quantifies the contribution of each component. We compare four configurations:
(1)\textbf{GRPO}: The baseline GRPO algorithm as employed in Search-R1.
(2)\textbf{GRPO w/ KL}: GRPO augmented with an unconditional KL penalty term.
(3)\textbf{GRPO w/ KL\_r}: GRPO with a KL penalty conditioned solely on the IS ratio.
(4)\textbf{GRPO w/ KL\_ra (SAPO)}: The complete SAPO configuration, where the KL penalty is conditioned on both the IS ratios and the advantage.

\paragraph{Impact of Selective KL Penalties.}
Table.\ref{tab_exp_ablation} details the performance progression, starting with the Search‑R1‑GRPO baseline, which achieves an average accuracy of 0.388. Incorporating a simple unconditional KL term yields a significant performance on Musique (+30.6\%) and HotpotQA (+14.4\%). This results in a modest increase in average accuracy to 0.398. The ratio-conditioned variant ($KL\_r$) further improves stability, raising the average to 0.417 and notably improving the HotpotQA score to 0.457. Finally, the full SAPO configuration ($KL\_ra$) achieves the best overall balance, securing the highest scores on six out of seven benchmarks with an average of 0.429, an absolute increase of 0.041 over the baseline (about 10.6\% relative). These findings validate that the conditional KL penalty in SAPO effectively balances exploration and constraint, leading to consistent aggregate improvements.

\definecolor{my-purple}{RGB}{112, 48, 160}
\definecolor{my-cyan}{RGB}{0, 176, 240}

\newcommand{\cinc}[1]{\multicolumn{1}{c}{\textcolor{my-purple}{\scriptsize (+#1)}}}
\newcommand{\cdec}[1]{\multicolumn{1}{c}{\textcolor{my-cyan}{\scriptsize (-#1)}}}
\newcommand{\cgray}[1]{\multicolumn{1}{c}{\textcolor{gray}{\scriptsize (#1)}}}

\begin{table*}[t]
    \centering
    \small
    \setlength{\tabcolsep}{3.6pt}
    \renewcommand{\arraystretch}{1.0}
    
    \begin{tabular}{lcccccccc}
        \toprule
        \multirow{2}{*}{\textbf{Method}} & \multicolumn{3}{c}{\textbf{Single-Hop QA}} & \multicolumn{4}{c}{\textbf{Multi-Hop QA}} &  \multirow{2}{*}{\textbf{Avg.}}\\
        \cmidrule(lr){2-4} \cmidrule(lr){5-8}
         & NQ & TriviaQA & PopQA & HotpotQA & 2Wiki & Musique & Bamboogle & \\
        \midrule
        
        Search-R1-GRPO 
        & 0.402 & 0.662 & 0.436 & 0.374 & 0.376 & 0.134 & 0.333 & 0.388 \\
        \addlinespace[0.2em] 
        
        \quad + $KL$ 
        & 0.425 & 0.625 & 0.443 & 0.428 & 0.371 & 0.175 & 0.342 & 0.398 \\
        & \cinc{0.023} & \cdec{0.037} & \cinc{0.007} 
        & \cinc{0.054} & \cdec{0.005} & \cinc{0.041} 
        & \cinc{0.009} & \cinc{0.010} \\
        \addlinespace[0.2em] 
        
        \quad + $KL_r$ 
        & 0.439 & 0.660 & 0.456 & \textbf{0.457} & 0.369 & 0.169 & 0.369 & 0.417 \\
        & \cinc{0.014} & \cinc{0.035} & \cinc{0.013} 
        & \cinc{0.029} & \cdec{0.002} & \cdec{0.006} 
        & \cinc{0.027} & \cinc{0.019} \\
        \addlinespace[0.2em] 
        
        \rowcolor{gray!10} 
        \quad + $KL_{ra}$ (SAPO) 
        & \textbf{0.442} & \textbf{0.673} & \textbf{0.474} & 0.433 & \textbf{0.415} & \textbf{0.181} & \textbf{0.382} & \textbf{0.429} \\
        & \cinc{0.003} & \cinc{0.013} & \cinc{0.018} 
        & \cdec{0.024} & \cinc{0.046} & \cinc{0.012} 
        & \cinc{0.013} & \cinc{0.012} \\
        
        \bottomrule
    \end{tabular}
    \caption{Ablation results on different KL terms in \textbf{SAPO}. The values in parentheses denote the absolute performance change compared to the previous row.}
    \label{tab_exp_ablation}
\end{table*}

\begin{figure*}[h]
	\centering
	\includegraphics[width=0.98\linewidth]{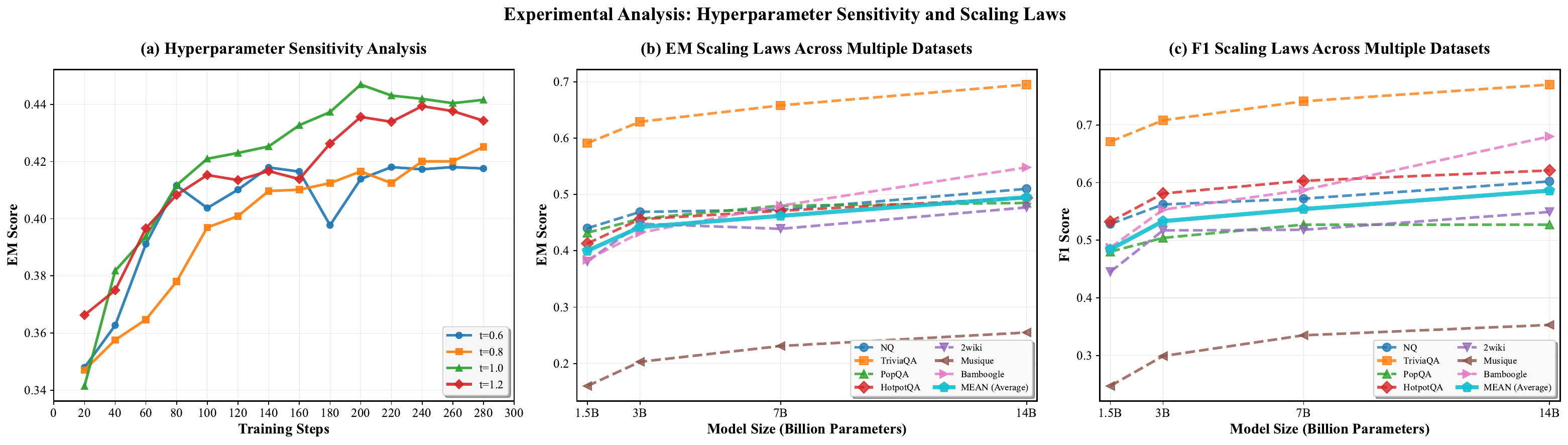}
	\caption{(a) Hyperparameter sensitivity analysis. (b,c) Scaling trends of Qwen2.5-Instruct with \textbf{SAPO} across different model sizes (1.5B to 14B). We report both EM and F1 scores.}
	\label{fig2}
\end{figure*}

\definecolor{headergray}{gray}{0.92}

\begin{table*}[t]
    \centering
    \small
    \setlength{\tabcolsep}{3.5pt} 
    \begin{tabular}{lcccccccc}
        \toprule
        \multirow{2}{*}{\textbf{Method}} & \multicolumn{3}{c}{\textbf{Single-Hop QA}} & \multicolumn{4}{c}{\textbf{Multi-Hop QA}} & \multirow{2}{*}{\textbf{Avg.}} \\
        \cmidrule(lr){2-4} \cmidrule(lr){5-8}
        &  NQ & TriviaQA & PopQA & HotpotQA & 2Wiki & Musique & Bamboogle  & \\
        \midrule
        
        \rowcolor{headergray} 
        \multicolumn{9}{l}{\textit{\textbf{Backbone: LLaMA-3.2-3B-Base}}} \\
        \addlinespace[0.2em]
        Search-R1-GRPO & 0.394 & 0.596 & 0.437 & 0.280 & 0.264 & 0.056 & 0.105 & 0.305 \\
        Search-R1-SAPO & \textbf{0.471} & \textbf{0.630} & \textbf{0.452} & \textbf{0.355} & \textbf{0.301} & \textbf{0.102} & \textbf{0.248} & \textbf{0.366} \\
        \addlinespace[0.5em]
        
        \rowcolor{headergray} 
        \multicolumn{9}{l}{\textit{\textbf{Backbone: LLaMA-3.2-3B-Instruct}}} \\
        \addlinespace[0.2em]
        Search-R1-GRPO & 0.357 & 0.578 & 0.378 & 0.314 & 0.233 & 0.090 & 0.306 & 0.322 \\
        Search-R1-SAPO & \textbf{0.475} & \textbf{0.634} & \textbf{0.446} & \textbf{0.444} & \textbf{0.386} & \textbf{0.187} & \textbf{0.384} & \textbf{0.422} \\
        
        \bottomrule
    \end{tabular}
    \caption{Generalization results on LLaMA-3.2-3B (Base and Instruct). \textbf{SAPO} consistently outperforms the GRPO baseline across both model variants.}
    \label{tab_generalization}
\end{table*}


\subsection{Detailed Analysis}

\paragraph{Hyperparameter Sensitivity Analysis.} We analyze the sensitivity of the IS threshold $t$ within our SAPO method. The validation EM scores in Figure~\ref{fig2}(a) indicate that $t = 1.0$ consistently yields the best performance throughout the training process. Specifically, at step 280, the model achieves an EM score of 0.442. In contrast, $t = 1.2$ results in a final score of 0.434, while $t = 0.6$ and $t = 0.8$ lead to inferior results of 0.418 and 0.425, respectively. Notably, the performance divergence among different threshold values widens as training progresses, suggesting that the selection of $t$ becomes increasingly critical in later stages. These results confirm that $t = 1.0$ strikes an optimal balance between exploration and exploitation, thereby being adopted as the default setting for all subsequent experiments.

\paragraph{Scaling with Model Size.} We investigate the scalability of SAPO by evaluating its performance across backbone models of varying sizes, specifically the Qwen2.5-Instruct series ranging from 1.5B to 14B parameters. Figure.~\ref{fig2} reports the EM and F1 scores. We observe a strong positive correlation between model size and performance, aligning with established scaling laws ~\cite{kaplan2020scalinglaw}. As the parameter count increases, SAPO demonstrates a monotonic improvement in average accuracy. Notably, scaling from 1.5B to 14B results in a substantial performance leap, raising the average EM from 0.400 to 0.495 and the F1 score from 0.484 to 0.586. This confirms that SAPO effectively leverages the enhanced reasoning capabilities inherent in LLM.

\paragraph{Robustness across Model Families.}
To assess the generalizability of SAPO across different architectures, we extend our evaluation to the LLaMA-3.2\cite{touvron2023llama} series, testing both the Base and Instruct versions of the 3B model. As detailed in Table \ref{tab_generalization}, SAPO consistently outperforms the GRPO baseline across both variants. On LLaMA-3.2-3B-Base, SAPO improves the average accuracy from 0.305 to 0.366, proving its capacity to align base pre-trained models effectively. The gains are even more pronounced on LLaMA-3.2-3B-Instruct, where SAPO boosts average accuracy by 10.0 percentage points (from 0.322 to 0.422). These results indicate that SAPO is a model-agnostic framework capable of enhancing search capabilities regardless of the underlying model family or its prior alignment stage.

\section{Conclusion}
Tool-based agentic reinforcement learning, such as GRPO, introduces Importance Sampling Distribution Drift (ISDD). This drift often causes the importance sampling ratio to decline sharply, leading to training instabilities that manifest as catastrophic and irreversible model collapse. In this paper, we propose SAPO (\textbf{S}earch \textbf{A}gent \textbf{P}olicy \textbf{O}ptimization) to enhance the training stability of search agents. We analyze the phenomena of increasing clipping rates and decreasing rewards, attributing these effects to ISDD. To mitigate this, SAPO introduces a penalty term that enforces a token-level conditonal constraint on the distributional divergence between the current and old policies. Extensive experiments on single-hop and multi-hop QA benchmarks demonstrate that SAPO significantly outperforms existing methods, improving absolute performance by $10.6$ points over the Search-R1 baseline.

\bibliography{colm2026_conference}

@article{kwiatkowski2019NQ,
  title={Natural questions: a benchmark for question answering research},
  author={Kwiatkowski, Tom and Palomaki, Jennimaria and Redfield, Olivia and Collins, Michael and Parikh, Ankur and Alberti, Chris and Epstein, Danielle and Polosukhin, Illia and Devlin, Jacob and Lee, Kenton and others},
  journal={Transactions of the Association for Computational Linguistics},
  volume={7},
  pages={453--466},
  year={2019},
  publisher={MIT Press One Rogers Street, Cambridge, MA 02142-1209, USA journals-info~…}
}

@article{mallen2022popqa,
  title={When not to trust language models: Investigating effectiveness of parametric and non-parametric memories},
  author={Mallen, Alex and Asai, Akari and Zhong, Victor and Das, Rajarshi and Khashabi, Daniel and Hajishirzi, Hannaneh},
  journal={arXiv preprint arXiv:2212.10511},
  year={2022}
}

@article{joshi2017triviaqa,
  title={Triviaqa: A large scale distantly supervised challenge dataset for reading comprehension},
  author={Joshi, Mandar and Choi, Eunsol and Weld, Daniel S and Zettlemoyer, Luke},
  journal={arXiv preprint arXiv:1705.03551},
  year={2017}
}

@article{yang2018hotpotqa,
  title={HotpotQA: A dataset for diverse, explainable multi-hop question answering},
  author={Yang, Zhilin and Qi, Peng and Zhang, Saizheng and Bengio, Yoshua and Cohen, William W and Salakhutdinov, Ruslan and Manning, Christopher D},
  journal={arXiv preprint arXiv:1809.09600},
  year={2018}
}

@article{ho2020wikimultihopqa,
  title={Constructing a multi-hop qa dataset for comprehensive evaluation of reasoning steps},
  author={Ho, Xanh and Nguyen, Anh-Khoa Duong and Sugawara, Saku and Aizawa, Akiko},
  journal={arXiv preprint arXiv:2011.01060},
  year={2020}
}

@article{trivedi2022musique,
  title={MuSiQue: Multihop Questions via Single-hop Question Composition},
  author={Trivedi, Harsh and Balasubramanian, Niranjan and Khot, Tushar and Sabharwal, Ashish},
  journal={Transactions of the Association for Computational Linguistics},
  volume={10},
  pages={539--554},
  year={2022},
  publisher={MIT Press One Broadway, 12th Floor, Cambridge, Massachusetts 02142, USA}
}

@article{press2022bamboogle,
  title={Measuring and narrowing the compositionality gap in language models},
  author={Press, Ofir and Zhang, Muru and Min, Sewon and Schmidt, Ludwig and Smith, Noah A and Lewis, Mike},
  journal={arXiv preprint arXiv:2210.03350},
  year={2022}
}

@article{jin2025searchr1,
  title={Search-r1: Training llms to reason and leverage search engines with reinforcement learning},
  author={Jin, Bowen and Zeng, Hansi and Yue, Zhenrui and Yoon, Jinsung and Arik, Sercan and Wang, Dong and Zamani, Hamed and Han, Jiawei},
  journal={arXiv preprint arXiv:2503.09516},
  year={2025}
}

@article{guo2025deepseekr1,
  title={Deepseek-r1: Incentivizing reasoning capability in llms via reinforcement learning},
  author={Guo, Daya and Yang, Dejian and Zhang, Haowei and Song, Junxiao and Zhang, Ruoyu and Xu, Runxin and Zhu, Qihao and Ma, Shirong and Wang, Peiyi and Bi, Xiao and others},
  journal={arXiv preprint arXiv:2501.12948},
  year={2025}
}

@article{lewis2020rag,
  title={Retrieval-augmented generation for knowledge-intensive nlp tasks},
  author={Lewis, Patrick and Perez, Ethan and Piktus, Aleksandra and Petroni, Fabio and Karpukhin, Vladimir and Goyal, Naman and K{\"u}ttler, Heinrich and Lewis, Mike and Yih, Wen-tau and Rockt{\"a}schel, Tim and others},
  journal={Advances in neural information processing systems},
  volume={33},
  pages={9459--9474},
  year={2020}
}

@article{li2025search-o1,
  title={Search-o1: Agentic search-enhanced large reasoning models},
  author={Li, Xiaoxi and Dong, Guanting and Jin, Jiajie and Zhang, Yuyao and Zhou, Yujia and Zhu, Yutao and Zhang, Peitian and Dou, Zhicheng},
  journal={arXiv preprint arXiv:2501.05366},
  year={2025}
}

@article{chen2025ReSearch,
  title={Learning to reason with search for llms via reinforcement learning},
  author={Chen, Mingyang and Sun, Linzhuang and Li, Tianpeng and Sun, Haoze and Zhou, Yijie and Zhu, Chenzheng and Wang, Haofen and Pan, Jeff Z and Zhang, Wen and Chen, Huajun and others},
  journal={arXiv preprint arXiv:2503.19470},
  year={2025}
}

@article{trivedi2022ircot,
  title={Interleaving retrieval with chain-of-thought reasoning for knowledge-intensive multi-step questions},
  author={Trivedi, Harsh and Balasubramanian, Niranjan and Khot, Tushar and Sabharwal, Ashish},
  journal={arXiv preprint arXiv:2212.10509},
  year={2022}
}

@inproceedings{karpukhin2020wikipediadump,
  title={Dense Passage Retrieval for Open-Domain Question Answering.},
  author={Karpukhin, Vladimir and Oguz, Barlas and Min, Sewon and Lewis, Patrick SH and Wu, Ledell and Edunov, Sergey and Chen, Danqi and Yih, Wen-tau},
  booktitle={EMNLP (1)},
  pages={6769--6781},
  year={2020}
}

@article{wang2022e5,
  title={Text embeddings by weakly-supervised contrastive pre-training},
  author={Wang, Liang and Yang, Nan and Huang, Xiaolong and Jiao, Binxing and Yang, Linjun and Jiang, Daxin and Majumder, Rangan and Wei, Furu},
  journal={arXiv preprint arXiv:2212.03533},
  year={2022}
}

@article{zhao2023surveyllm,
  title={A survey of large language models},
  author={Zhao, Wayne Xin and Zhou, Kun and Li, Junyi and Tang, Tianyi and Wang, Xiaolei and Hou, Yupeng and Min, Yingqian and Zhang, Beichen and Zhang, Junjie and Dong, Zican and others},
  journal={arXiv preprint arXiv:2303.18223},
  volume={1},
  number={2},
  year={2023}
}

@article{touvron2023llama,
  title={Llama: Open and efficient foundation language models},
  author={Touvron, Hugo and Lavril, Thibaut and Izacard, Gautier and Martinet, Xavier and Lachaux, Marie-Anne and Lacroix, Timoth{\'e}e and Rozi{\`e}re, Baptiste and Goyal, Naman and Hambro, Eric and Azhar, Faisal and others},
  journal={arXiv preprint arXiv:2302.13971},
  year={2023}
}

@article{kaelbling1996rl,
  title={Reinforcement learning: A survey},
  author={Kaelbling, Leslie Pack and Littman, Michael L and Moore, Andrew W},
  journal={Journal of artificial intelligence research},
  volume={4},
  pages={237--285},
  year={1996}
}

@article{shao2024deepseekmath,
  title={Deepseekmath: Pushing the limits of mathematical reasoning in open language models},
  author={Shao, Zhihong and Wang, Peiyi and Zhu, Qihao and Xu, Runxin and Song, Junxiao and Bi, Xiao and Zhang, Haowei and Zhang, Mingchuan and Li, YK and Wu, Yang and others},
  journal={arXiv preprint arXiv:2402.03300},
  year={2024}
}

@article{shi2025autorefine,
  title={Search and refine during think: Autonomous retrieval-augmented reasoning of llms},
  author={Shi, Yaorui and Li, Shihan and Wu, Chang and Liu, Zhiyuan and Fang, Junfeng and Cai, Hengxing and Zhang, An and Wang, Xiang},
  journal={arXiv e-prints},
  pages={arXiv--2505},
  year={2025}
}

@article{kobayashi2000ir,
  title={Information retrieval on the web},
  author={Kobayashi, Mei and Takeda, Koichi},
  journal={ACM computing surveys (CSUR)},
  volume={32},
  number={2},
  pages={144--173},
  year={2000},
  publisher={ACM New York, NY, USA}
}

@article{li2025webthinker,
  title={Webthinker: Empowering large reasoning models with deep research capability},
  author={Li, Xiaoxi and Jin, Jiajie and Dong, Guanting and Qian, Hongjin and Wu, Yongkang and Wen, Ji-Rong and Zhu, Yutao and Dou, Zhicheng},
  journal={arXiv preprint arXiv:2504.21776},
  year={2025}
}

@article{wu2025webdancer,
  title={Webdancer: Towards autonomous information seeking agency},
  author={Wu, Jialong and Li, Baixuan and Fang, Runnan and Yin, Wenbiao and Zhang, Liwen and Tao, Zhengwei and Zhang, Dingchu and Xi, Zekun and Fu, Gang and Jiang, Yong and others},
  journal={arXiv preprint arXiv:2505.22648},
  year={2025}
}

@article{li2025websailor,
  title={WebSailor: Navigating Super-human Reasoning for Web Agent},
  author={Li, Kuan and Zhang, Zhongwang and Yin, Huifeng and Zhang, Liwen and Ou, Litu and Wu, Jialong and Yin, Wenbiao and Li, Baixuan and Tao, Zhengwei and Wang, Xinyu and others},
  journal={arXiv preprint arXiv:2507.02592},
  year={2025}
}

@article{mei20252o2searcher,
  title={$O^{2}$-Searcher: A Searching-based Agent Model for Open-Domain Open-Ended Question Answering},
  author={Mei, Jianbiao and Hu, Tao and Fu, Daocheng and Wen, Licheng and Yang, Xuemeng and Wu, Rong and Cai, Pinlong and Cai, Xinyu and Gao, Xing and Yang, Yu and others},
  journal={arXiv preprint arXiv:2505.16582},
  year={2025}
}

@article{qian2025inforage,
  title={Scent of Knowledge: Optimizing Search-Enhanced Reasoning with Information Foraging},
  author={Qian, Hongjin and Liu, Zheng},
  journal={arXiv preprint arXiv:2505.09316},
  year={2025}
}

@article{schulman2017ppo,
  title={Proximal policy optimization algorithms},
  author={Schulman, John and Wolski, Filip and Dhariwal, Prafulla and Radford, Alec and Klimov, Oleg},
  journal={arXiv preprint arXiv:1707.06347},
  year={2017}
}

@article{rafailov2023dpo,
  title={Direct preference optimization: Your language model is secretly a reward model},
  author={Rafailov, Rafael and Sharma, Archit and Mitchell, Eric and Manning, Christopher D and Ermon, Stefano and Finn, Chelsea},
  journal={Advances in neural information processing systems},
  volume={36},
  pages={53728--53741},
  year={2023}
}

@article{yu2025dapo,
  title={Dapo: An open-source llm reinforcement learning system at scale},
  author={Yu, Qiying and Zhang, Zheng and Zhu, Ruofei and Yuan, Yufeng and Zuo, Xiaochen and Yue, Yu and Dai, Weinan and Fan, Tiantian and Liu, Gaohong and Liu, Lingjun and others},
  journal={arXiv preprint arXiv:2503.14476},
  year={2025}
}

@article{zheng2025gspo,
  title={Group sequence policy optimization},
  author={Zheng, Chujie and Liu, Shixuan and Li, Mingze and Chen, Xiong-Hui and Yu, Bowen and Gao, Chang and Dang, Kai and Liu, Yuqiong and Men, Rui and Yang, An and others},
  journal={arXiv preprint arXiv:2507.18071},
  year={2025}
}

@misc{qwen2025qwen25technicalreport,
      title={Qwen2.5 Technical Report}, 
      author={Qwen and : and An Yang and Baosong Yang and Beichen Zhang and Binyuan Hui and Bo Zheng and Bowen Yu and Chengyuan Li and Dayiheng Liu and Fei Huang and Haoran Wei and Huan Lin and Jian Yang and Jianhong Tu and Jianwei Zhang and Jianxin Yang and Jiaxi Yang and Jingren Zhou and Junyang Lin and Kai Dang and Keming Lu and Keqin Bao and Kexin Yang and Le Yu and Mei Li and Mingfeng Xue and Pei Zhang and Qin Zhu and Rui Men and Runji Lin and Tianhao Li and Tianyi Tang and Tingyu Xia and Xingzhang Ren and Xuancheng Ren and Yang Fan and Yang Su and Yichang Zhang and Yu Wan and Yuqiong Liu and Zeyu Cui and Zhenru Zhang and Zihan Qiu},
      year={2025},
      eprint={2412.15115},
      archivePrefix={arXiv},
      primaryClass={cs.CL},
      url={https://arxiv.org/abs/2412.15115}, 
}

@inproceedings{sheng2025hybridflow,
  title={Hybridflow: A flexible and efficient rlhf framework},
  author={Sheng, Guangming and Zhang, Chi and Ye, Zilingfeng and Wu, Xibin and Zhang, Wang and Zhang, Ru and Peng, Yanghua and Lin, Haibin and Wu, Chuan},
  booktitle={Proceedings of the Twentieth European Conference on Computer Systems},
  pages={1279--1297},
  year={2025}
}

@article{guo2024agentsurvey,
  title={Large language model based multi-agents: A survey of progress and challenges},
  author={Guo, Taicheng and Chen, Xiuying and Wang, Yaqi and Chang, Ruidi and Pei, Shichao and Chawla, Nitesh V and Wiest, Olaf and Zhang, Xiangliang},
  journal={arXiv preprint arXiv:2402.01680},
  year={2024}
}

@article{li2025aissurvey,
  title={  },
  author={Li, Jian and Li, Xiaoxi and Zheng, Yan and Jin, Yizhang and Wang, Shuo and Wu, Jiafu and Wang, Yabiao and Wang, Chengjie and Yuan, Xiaotong},
  year={2025}
}

@article{wang2025stepsearch,
  title={StepSearch: Igniting LLMs Search Ability via Step-Wise Proximal Policy Optimization},
  author={Wang, Ziliang and Zheng, Xuhui and An, Kang and Ouyang, Cijun and Cai, Jialu and Wang, Yuhang and Wu, Yichao},
  journal={arXiv preprint arXiv:2505.15107},
  year={2025}
}

@article{sun2025zerosearch,
  title={Zerosearch: Incentivize the search capability of llms without searching},
  author={Sun, Hao and Qiao, Zile and Guo, Jiayan and Fan, Xuanbo and Hou, Yingyan and Jiang, Yong and Xie, Pengjun and Zhang, Yan and Huang, Fei and Zhou, Jingren},
  journal={arXiv preprint arXiv:2505.04588},
  year={2025}
}

@article{zhang2025criticsearch,
  title={CriticSearch: Fine-Grained Credit Assignment for Search Agents via a Retrospective Critic},
  author={Zhang, Yaocheng and Huang, Haohuan and Song, Zijun and Zhu, Yuanheng and Zhang, Qichao and Zhao, Zijie and Zhao, Dongbin},
  journal={arXiv preprint arXiv:2511.12159},
  year={2025}
}

@article{kaplan2020scalinglaw,
  title={Scaling laws for neural language models},
  author={Kaplan, Jared and McCandlish, Sam and Henighan, Tom and Brown, Tom B and Chess, Benjamin and Child, Rewon and Gray, Scott and Radford, Alec and Wu, Jeffrey and Amodei, Dario},
  journal={arXiv preprint arXiv:2001.08361},
  year={2020}
}

@article{DAPO,
  title   = {DAPO: An Open-Source LLM Reinforcement Learning System at Scale},
  author  = {Yu, Qiying and Zhang, Zheng and Zhu, Ruofei and Yuan, Yufeng and Zuo, Xiaochen and Yue, Yu and Dai, Weinan and Fan, Tiantian and Liu, Gaohong and Liu, Lingjun and others},
  journal = {arXiv preprint arXiv:2503.14476},
  year    = {2025}
}

@article{SimpleRL-Zoo,
  title   = {SimpleRL-Zoo: Investigating and Taming Zero Reinforcement Learning for Open Base Models in the Wild},
  author  = {Zeng, Weihao and Huang, Yuzhen and Liu, Qian and Liu, Wei and He, Keqing and Ma, Zejun and He, Junxian},
  journal = {arXiv preprint arXiv:2503.18892},
  year    = {2025}
}

@article{Skywork-OR1,
  title   = {Skywork Open Reasoner 1 Technical Report},
  author  = {He, Jujie and Liu, Jiacai and Liu, Chris Yuhao and Yan, Rui and Wang, Chaojie and Cheng, Peng and Zhang, Xiaoyu and Zhang, Fuxiang and Xu, Jiacheng and Shen, Wei and others},
  journal = {arXiv preprint arXiv:2505.22312},
  year    = {2025}
}

@article{CISPO,
  title   = {MiniMax-M1: Scaling Test-Time Compute Efficiently with Lightning Attention},
  author  = {Chen, Aili and Li, Aonian and Gong, Bangwei and Jiang, Binyang and Fei, Bo and Yang, Bo and Shan, Boji and Yu, Changqing and Wang, Chao and Zhu, Cheng and others},
  journal = {arXiv preprint arXiv:2506.13585},
  year    = {2025}
}

@misc{POLARIS,
  title  = {POLARIS: A Post-Training Recipe for Scaling Reinforcement Learning on Advanced Reasoning Models},
  url    = {https://hkunlp.github.io/blog/2025/Polaris},
  author = {An, Chenxin and Xie, Zhihui and Li, Xiaonan and Li, Lei and Zhang, Jun and Gong, Shansan and Zhong, Ming and Xu, Jingjing and Qiu, Xipeng and Wang, Mingxuan and Kong, Lingpeng},
  year   = {2025}
}

@article{Archer,
  title   = {Stabilizing Knowledge, Promoting Reasoning: Dual-Token Constraints for RLVR},
  author  = {Wang, Jiakang and Liu, Runze and Zhang, Fuzheng and Li, Xiu and Zhou, Guorui},
  journal = {arXiv preprint arXiv:2507.15778},
  year    = {2025}
}

@article{zhang2025survey,
  title   = {A Survey of Reinforcement Learning for Large Reasoning Models},
  author  = {Zhang, Kaiyan and Zuo, Yuxin and He, Bingxiang and Sun, Youbang and Liu, Runze and Jiang, Che and Fan, Yuchen and Tian, Kai and Jia, Guoli and Li, Pengfei and Fu, Yu and Lv, Xingtai and Zhang, Yuchen and Zeng, Sihang and Qu, Shang and Li, Haozhan and Wang, Shijie and Wang, Yuru and Long, Xinwei and Liu, Fangfu and Xu, Xiang and Ma, Jiaze and Zhu, Xuekai and Hua, Ermo and Liu, Yihao and Li, Zonglin and Chen, Huayu and Qu, Xiaoye and Li, Yafu and Chen, Weize and Yuan, Zhenzhao and Gao, Junqi and Li, Dong and Ma, Zhiyuan and Cui, Ganqu and Liu, Zhiyuan and Qi, Biqing and Ding, Ning and Zhou, Bowen},
  journal = {arXiv preprint arXiv:2509.08827},
  year    = {2025}
}

@article{shi2025exsearch,
  title={Iterative self-incentivization empowers large language models as agentic searchers},
  author={Shi, Zhengliang and Yan, Lingyong and Yin, Dawei and Verberne, Suzan and de Rijke, Maarten and Ren, Zhaochun},
  journal={arXiv preprint arXiv:2505.20128},
  year={2025}
}

@article{deng2025nthr,
  title={On the effect of negative gradient in group relative deep reinforcement optimization},
  author={Deng, Wenlong and Ren, Yi and Li, Muchen and Sutherland, Danica J and Li, Xiaoxiao and Thrampoulidis, Christos},
  journal={arXiv preprint arXiv:2505.18830},
  year={2025}
}

@article{deng2025llds,
  title={On grpo collapse in search-r1: The lazy likelihood-displacement death spiral},
  author={Deng, Wenlong and Li, Yushu and Gong, Boying and Ren, Yi and Thrampoulidis, Christos and Li, Xiaoxiao},
  journal={arXiv preprint arXiv:2512.04220},
  year={2025}
}

@article{wang2026arlarena,
  title={ARLArena: A Unified Framework for Stable Agentic Reinforcement Learning},
  author={Wang, Xiaoxuan and Zhang, Han and Wang, Haixin and Shi, Yidan and Li, Ruoyan and Han, Kaiqiao and Tong, Chenyi and Deng, Haoran and Sun, Renliang and Taylor, Alexander and others},
  journal={arXiv preprint arXiv:2602.21534},
  year={2026}
}

@misc{li2026sesearch,
      title={SE-Search: Self-Evolving Search Agent via Memory and Dense Reward}, 
      author={Jian Li and Yizhang Jin and Dongqi Liu and Hang Ding and Jiafu Wu and Dongsheng Chen and Yunhang Shen and Yulei Qin and Ying Tai and Chengjie Wang and Xiaotong Yuan and Yabiao Wang},
      year={2026},
      eprint={2603.03293},
      archivePrefix={arXiv},
      primaryClass={cs.CL},
      url={https://arxiv.org/abs/2603.03293}, 
}
\bibliographystyle{colm2026_conference}

\newpage
\appendix
\section{Appendix}

\subsection{Related Works}

\paragraph{Retrieval-Augmented Generation}
Large Language Models (LLMs)~\cite{zhao2023surveyllm} have demonstrated remarkable capabilities in natural language understanding, reasoning, and information synthesis. However, they lack inherent access to real-time external knowledge and remain susceptible to hallucinations and factual errors. Retrieval-Augmented Generation (RAG)~\cite{lewis2020rag} addresses these limitations by integrating LLMs with retrieval engines, conditioning generation on retrieved external passages to mitigate inaccuracies. Despite their utility, static RAG pipelines typically employ a ``retrieve-then-generate'' paradigm, which restricts the LLM's autonomy in determining \textit{when} and \textit{what} to search. This rigidity limits their adaptability in complex, real-world applications where iterative information seeking is required.

\paragraph{Search Agents}
While traditional search engines (e.g., Google, Bing) facilitate efficient access to ranked web pages, they often struggle to capture nuanced user intent in complex queries~\cite{kobayashi2000ir}. Search agents \cite{li2025aissurvey,li2026sesearch} represent an evolution beyond static RAG by invoking search tools within a multi-step reasoning process, allowing for autonomous query refinement and iterative retrieval. Systems such as Search-o1~\cite{li2025search-o1}, WebThinker~\cite{li2025webthinker}, and WebDancer~\cite{wu2025webdancer} exemplify this capability. Concurrently, research has focused on training strategies to optimize search behaviors, including Search-R1~\cite{jin2025searchr1}, ReSearch~\cite{chen2025ReSearch}, AutoRefine~\cite{shi2025autorefine}, SE-Search ~\cite{li2026sesearch} and WebSailor~\cite{li2025websailor}. However, a pervasive limitation among these methods is post-training instability, which frequently manifests as catastrophic model collapse or irreversible performance degradation. To mitigate this, we propose Search Agent Policy Optimization (SAPO), a method designed to enhance training stability through constrained policy updates.

\paragraph{Tool-based Agentic Reinforcement Learning}
Reinforcement Learning (RL) provides a robust framework for aligning LLM behaviors with human intent. Prominent approaches include Proximal Policy Optimization (PPO)~\cite{schulman2017ppo}, Direct Preference Optimization (DPO)~\cite{rafailov2023dpo}, and Group Relative Policy Optimization (GRPO)~\cite{shao2024deepseekmath}. GRPO is particularly notable for its use of group-based normalization, which eliminates the need for a separate value network. Extensions such as Dynamic Sampling Policy Optimization (DAPO)~\cite{yu2025dapo} and Group Sequence Policy Optimization (GSPO)~\cite{zheng2025gspo} have further refined this paradigm. Other group-based optimization methods include those detailed in~\cite{SimpleRL-Zoo, Skywork-OR1, CISPO, POLARIS, Archer, zhang2025survey}. Nevertheless, applying RL to tool-augmented agentic environments remains highly unstable due to the interactive, multi-turn nature of the tasks~\cite{wang2026arlarena, deng2025llds, deng2025nthr}. Addressing this challenge, our work introduces a sample-based penalty term to enforce a token-level constraint between the updated and reference policies, thereby stabilizing the training of search agents.

\subsection{Training and Evaluation Datasets}

Consistent with Search-R1 and other RL-based search agents, we construct our training dataset by aggregating NQ~\cite{kwiatkowski2019NQ} and HotpotQA~\cite{yang2018hotpotqa}, yielding a total of 169,615 samples. For evaluation, we construct a comprehensive benchmark of 51,713 samples, incorporating the test sets of four datasets (NQ~\cite{kwiatkowski2019NQ}, TriviaQA~\cite{joshi2017triviaqa}, PopQA~\cite{mallen2022popqa}, Bamboogle~\cite{press2022bamboogle}) and the development sets of three multi-hop datasets (HotpotQA~\cite{yang2018hotpotqa}, 2WikiMultihopQA~\cite{ho2020wikimultihopqa}, and Musique ~\cite{trivedi2022musique}).

\subsection{Search Agent Prompt}

\newcommand{\token}[1]{\texttt{\small\textless#1\textgreater}}

\definecolor{thinkcolor}{RGB}{0, 119, 187}   
\definecolor{searchcolor}{RGB}{0, 153, 136}  
\definecolor{doccolor}{RGB}{100, 100, 100}   
\definecolor{anscolor}{RGB}{204, 51, 17}     
\definecolor{tagcolor}{RGB}{238, 119, 51}    

\newcommand{\tagthink}[1]{\textcolor{thinkcolor}{\texttt{<think>}} #1 \textcolor{thinkcolor}{\texttt{</think>}}}
\newcommand{\tagsearch}[1]{\textcolor{searchcolor}{\texttt{<search>}} \textbf{#1} \textcolor{searchcolor}{\texttt{</search>}}}
\newcommand{\tagdoc}[1]{\textcolor{tagcolor}{\texttt{<documents>}} #1 \textcolor{tagcolor}{\texttt{</documents>}}}
\newcommand{\tagans}[1]{\textcolor{anscolor}{\texttt{<answer>}} \textbf{#1} \textcolor{anscolor}{\texttt{</answer>}}}

\begin{figure*}[t]
    \centering
    \small
    \vspace{-.5em}
    \begin{tcolorbox}[colback=gray!5!white, colframe=black!15, width=\linewidth, boxrule=0.5pt, arc=2mm]
    You are a helpful assistant excel at answering questions with multi-turn search engine calling. To answer questions, you must first reason through the available information using \token{think} and \token{/think}. If you identify missing knowledge, you may issue a search request using \token{search} query \token{/search} at any time. The retrieval system will provide you with the three most relevant documents enclosed in \token{documents} and \token{/documents}. You may send multiple search requests if needed. Once you have sufficient information, provide a concise final answer using \token{answer} and \token{/answer}. For example, \token{answer} Donald Trump \token{/answer}. Question: \{\textbf{question}\} 
    \end{tcolorbox}
    \vspace{-.5em}
    \caption{Prompt template for SAPO.}
    \label{fig_prompt_template}
\end{figure*}

Figure \ref{fig_prompt_template} illustrates the prompt template used for the SAPO search agent. The template instructs the assistant to perform multi-turn search-driven reasoning by alternating between internal thought processes in \tagthink{} and query generation in \tagsearch{}. Retrieved results are returned as up to three documents within \tagdoc{}. The design permits multiple search calls and directs the agent to consolidate evidence into a concise final response enclosed by \tagans{}.

\subsection{SAPO Hyperparameters}

We implement our method using the VeRL framework \cite{sheng2025hybridflow}. As detailed in Table~\ref{tab:hyperparams}, we maintain consistency with Search-R1 and SE-Search regarding data, actor, and rollout configurations.Specifically for SAPO, the KL penalty coefficient is set to $\gamma=0.1$, and the IS ratios threshold is set to $\tau=1.0$.

\begin{table}[htbp]
\centering
\caption{Hyperparameters used in SAPO.}
\label{tab:hyperparams}
\small
\begin{adjustbox}{width=0.45\textwidth}
\begin{tabular}{@{}l l c@{}}
\toprule
Module & Hyper-parameter & Value \\
\midrule
\addlinespace[0.5ex]
\textbf{Data} & Max documents length & 512 \\
             & Max response length  & 2048 \\
             & Total training steps & 420 \\
             & Retriever top-k      & 3 \\
\midrule
\textbf{Actor} & Training batch size      & 256 \\
               & Micro batch size         & 64 \\
               & Learning rate            & $1\times10^{-6}$ \\
               & KL coefficient $\beta$   & 0.001 \\
               & Clip ratio $\epsilon$    & 0.2 \\
\midrule
\textbf{Rollout} & Max search actions & 5 \\
                 & Group size $G$     & 10 \\
                 & Temperature        & 1.0 \\
                 & Top-p              & 0.95 \\
\midrule
\textbf{KL term} & KL penalty coeff \ $\gamma$ & 0.1 \\
                      & IS ratios threshold $\tau$   & 1.0 \\
\bottomrule
\end{tabular}
\end{adjustbox}
\end{table}





\subsection{Generalization and Scalability}
We investigate the scalability of SAPO by evaluating its performance across backbone models of varying capacities, specifically the Qwen2.5-Instruct series ranging from 1.5B to 14B parameters. To isolate the impact of model scale, we keep the retriever, knowledge corpus, and evaluation protocols constant. Table~\ref{tab_scaling} reports the Exact Match (EM) and F1 scores. The results exhibit a clear positive correlation between model size and performance, aligning with established scaling laws~\cite{kaplan2020scalinglaw}. 

\begin{table*}[t]
    \centering
    \small
    \setlength{\tabcolsep}{3.5pt}
    \begin{tabular}{llcccccccc}
        \toprule
        \multirow{2}{*}{\textbf{Size}} & \multirow{2}{*}{\textbf{Metric}} & \multicolumn{3}{c}{\textbf{Single-Hop QA}} & \multicolumn{4}{c}{\textbf{Multi-Hop QA}} & \multirow{2}{*}{\textbf{Avg.}} \\
        \cmidrule(lr){3-5} \cmidrule(lr){6-9}
        & &  NQ & TriviaQA & PopQA & HotpotQA & 2Wiki & Musique & Bamboogle  & \\
        \midrule
        
        \multirow{2}{*}{1.5B} 
        & EM & 0.440 & 0.591 & 0.432 & 0.413 & 0.381 & 0.160 & 0.384 & 0.400 \\
        & F1 & 0.528 & 0.671 & 0.480 & 0.532 & 0.445 & 0.247 & 0.487 & 0.484 \\
        \cmidrule(lr){1-10} 
        
        \multirow{2}{*}{3B} 
        & EM & 0.469 & 0.629 & 0.457 & 0.456 & 0.448 & 0.203 & 0.432 & 0.442 \\
        & F1 & 0.562 & 0.708 & 0.504 & 0.581 & 0.517 & 0.299 & 0.553 & 0.533 \\
        \cmidrule(lr){1-10}
        
        \multirow{2}{*}{7B} 
        & EM & 0.473 & 0.658 & 0.480 & 0.471 & 0.439 & 0.231 & 0.480 & 0.462 \\
        & F1 & 0.572 & 0.741 & 0.527 & 0.603 & 0.518 & 0.335 & 0.587 & 0.554 \\
        \cmidrule(lr){1-10}
        
        \multirow{2}{*}{14B} 
        & EM & \textbf{0.510} & \textbf{0.695} & \textbf{0.485} & \textbf{0.494} & \textbf{0.477} & \textbf{0.255} & \textbf{0.548} & \textbf{0.495} \\
        & F1 & \textbf{0.602} & \textbf{0.770} & \textbf{0.527} & \textbf{0.621} & \textbf{0.549} & \textbf{0.353} & \textbf{0.680} & \textbf{0.586} \\
        
        \bottomrule
    \end{tabular}
    \caption{Scaling trends of Qwen2.5-Instruct with \textbf{SAPO} across different model sizes (1.5B to 14B). We report both EM and F1 scores. The best results are marked in bold.}
    \label{tab_scaling}
\end{table*}

\subsection{Implementation Simplicity}

We augment the standard GRPO surrogate objective with a concise penalty term to enforce token-level constraints on distributional shifts between the current and old policies. As shown in Figure \ref{lst:compute_policy_loss}, the core contribution of SAPO is encapsulated in the \texttt{kl\_term\_loss} expression. This term yields a scalar, token-level divergence penalty that is added to the global loss to directly limit per-token policy drift. By operating at the token granularity and requiring only a single line of code, this modification provides a lightweight and interpretable regularizer that complements the existing PPO clipping mechanism while enabling finer control over distributional changes during policy updates.

\definecolor{codegray}{rgb}{0.5,0.5,0.5}
\definecolor{codebg}{rgb}{0.97,0.97,0.97}
\lstdefinestyle{pythonstyle}{
  language=Python,
  basicstyle=\ttfamily\small,
  keywordstyle=\color{blue}\bfseries,
  commentstyle=\color{codegray}\itshape,
  stringstyle=\color{red},
  backgroundcolor=\color{codebg},
  numbers=left,
  numberstyle=\tiny\color{codegray},
  numbersep=6pt,
  frame=single,
  xleftmargin=2.5em,
  xrightmargin=2.5em,
  breaklines=true,
  columns=fullflexible,
  showstringspaces=false
  moredelim=[is][\bfseries]{@}{@},
  moredelim=[is][\color{purple}]{!}{!}
}

\begin{lstlisting}[style=pythonstyle,label={lst:compute_policy_loss}]
 def compute_policy_loss(old_log_prob, log_prob, advantages, eos_mask, cliprange):
    negative_approx_kl = log_prob - old_log_prob
    ratio = torch.exp(negative_approx_kl)
    ppo_kl = verl_F.masked_mean(-negative_approx_kl, eos_mask)

    pg_losses = -advantages * ratio
    pg_losses2 = -advantages * torch.clamp(ratio, 1.0 - cliprange, 1.0 + cliprange)

    !kl_term_loss = -verl_F.masked_mean(
        torch.log(ratio).masked_fill(ratio > t, 0)[advantages[:, 0] >= 0],
        eos_mask[advantages[:, 0] >= 0]
    )!

    pg_loss = verl_F.masked_mean(torch.max(pg_losses, pg_losses2), eos_mask)
    pg_clipfrac = verl_F.masked_mean(torch.gt(pg_losses2, pg_losses).float(), eos_mask)
    return pg_loss + w * kl_term_loss, pg_clipfrac, ppo_kl
\end{lstlisting}

\subsection{Additional Experiments Results}
\begin{figure*}[h]
	\centering
	\includegraphics[width=0.98\linewidth]{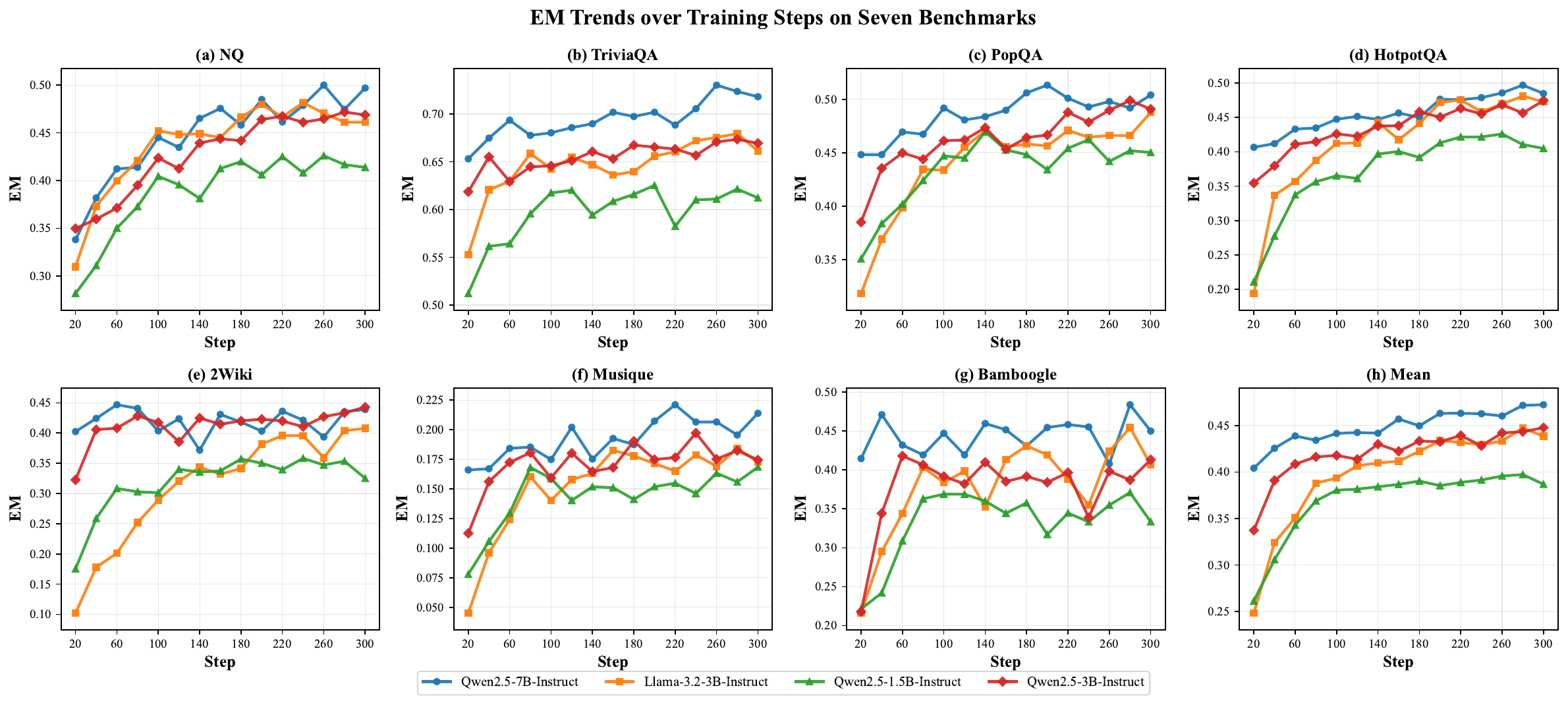}
	\caption{Evolution of EM accuracy over training steps for SAPO across seven benchmarks.}
	\label{fig_em7}
    \vspace{-1.em}
\end{figure*}

To provide a comprehensive characterization of SAPO's training stability, Figure \ref{fig_em7} illustrates the SAPO EM evolution of four backbones(Qwen2.5-7B-Instruct; Llama-3.2-3B-Instruct; Qwen2.5-1.5B-Instruct; Qwen2.5-3B-Instruct) across seven benchmarks: Natural Questions (NQ), TriviaQA, PopQA, Bamboogle, HotpotQA, 2Wiki, and Musique.

\subsection{Case Studies}
\label{app_case_study}

Table~\ref{tab_case1} and ~\ref{tab_case2} present full trajectories generated by the 3B and 14B models using SAPO, based on two randomly selected examples from 2WikiMultihopQA and Musique. These qualitative results demonstrate the agent's capacity for autonomous reasoning and iterative query refinement.

\begin{table*}[t!]
    \centering
    \small
    \caption{Case Studies illustrating SAPO  trajectories from the 2wikimultihopqa benchmark. Key evidence is extracted from the original searched documents.}
    \label{tab_case1}
    \vspace{-0.5em}
    
    \begin{tabularx}{\linewidth}{X} 
    \toprule
    \textbf{Question:} Which film has the director who died earlier, Deuces Wild or Cavalcade Of The West?\\
    \textbf{Ground Truth:} ["Cavalcade of the West", "Cavalcade Of The West"] \\
    \midrule
    \textbf{Model Output: Qwen2.5-3B-Instruct-SAPO} \\
    \midrule
    
    To determine which film has the director who died earlier between "Deuces Wild" and "Cavalcade Of The West", I need to gather information about the directors of these films and their death dates. Let's start by searching for this information.
    \vspace{0.4em} \\
    
    \tagsearch{Director of Deuces Wild} 
    \vspace{0.2em} \\
    \tagdoc{
        \textbf{Doc 1 (Title: "Deuces Wild"):} Deuces Wild is a 2002 American crime drama film directed by \textbf{Scott Kalvert}...\\
        \textbf{Doc 3 (Title: "Scott Kalvert"):} Scott Kalvert (August 15, 1964 -- \textbf{March 5, 2014}) was an American film director, known mainly for his 1995 film "The Basketball Diaries"... and 2002's "Deuces Wild".
    }
    \vspace{0.6em} \\
    \tagsearch{Director of Cavalcade Of The West}
    \vspace{0.2em} \\
    \tagdoc{
        \textbf{Doc 1 (Title: "Cavalcade of the West"):} Cavalcade of the West is a 1936 American Western film directed by \textbf{Harry L. Fraser}...\\
        \textbf{Doc 3 (Title: "Cavalcade of the West"):} ...Cavalcade of the West is a 1936 American Western film directed by Harry L. Fraser.
    }
    \vspace{0.6em} \\
    
    \tagsearch{Harry L. Fraser}
    \vspace{0.2em} \\
    \tagdoc{
        \textbf{Doc 1 (Title: "Harry W. Fraser"):} Harry W. Fraser (7 June 1884 -- \textbf{13 May 1950}) was an American labor leader... \\
        \textbf{Doc 3 (Title: "Harry W. Fraser"):} Fraser was president of the Order of Railway Conductors of America from 1941 to 1950. ... Fraser was born on June 7, 1884 and \textbf{died on May 13, 1950}.
    }
    \vspace{0.6em} \\
    \tagsearch{Scott Kalvert}
    \vspace{0.2em} \\
    \tagdoc{
        \textbf{Doc 1 (Title: "Scott Kalvert"):} Scott Kalvert (August 15, 1964 -- \textbf{March 5, 2014}) was an American film director... \\
        \textbf{Doc 2 (Title: "Scott Kalvert"):} ...on March 5, 2014 from an apparent suicide.
    }
    \vspace{0.6em} \\
    Based on the information provided, Scott Kalvert died on March 5, 2014, while Harry L. Fraser died on May 13, 1950. Therefore, the director who died earlier is Harry L. Fraser, and the film "Cavalcade Of The West" was directed by him.
    \vspace{0.4em} \\
    
    \tagans{Cavalcade Of The West} \\
    
    \bottomrule
    \end{tabularx}
\end{table*}

\newpage
\begin{table*}[t!] 
    \centering
    \small 
    \caption{Case Study illustrating SAPO from the Musique benchmark. Key evidence is extracted from the original searched documents.}
    \label{tab_case2}
    \vspace{-0.5em}
    
    \begin{tabularx}{\linewidth}{X} 
    \toprule
    \textbf{Question:} Who is the president of the newly declared independent country that is part of the Commission of Truth and Friendship with the country that Tony Gunawan is from?\\
    \textbf{Ground Truth:} ["Francisco Guterres"] \\
    \midrule
    \textbf{Model Output: Qwen2.5-14B-Instruct-SAPO} \\
    \midrule
    
    \tagthink{To answer this question, I need to identify the country that Tony Gunawan is from, find out if it has a Commission of Truth and Friendship with any newly declared independent country, and then find out who the president of that newly declared independent country is.}
    \vspace{0.4em} \\
    
    \tagsearch{Tony Gunawan country of origin} 
    \vspace{0.2em} \\
    
    \tagdoc{
        \textbf{Doc 1 (Title: "Tony Gunawan"):} Tony Gunawan (, born 9 April 1975) is a Chinese descended, \textbf{Indonesian born} badminton player... Gunawan is a former Olympic gold medalist and world champion for Indonesia...\\
        \textbf{Doc 2 (Title: "Tony Gunawan"):} ...Tony played for Indonesia from 1992–2001...\\
        \textbf{Doc 3 (Title: "Tony Gunawan"):} ...He was a member of the world champion 2000 Indonesia Thomas Cup team...
    }
    \vspace{0.6em} \\

    \tagthink{From the documents, it is clear that Tony Gunawan is from Indonesia. Now I need to find out if Indonesia has a Commission of Truth and Friendship with any newly declared independent country.}
    \vspace{0.4em} \\
    
    \tagsearch{Commission of Truth and Friendship Indonesia}
    \vspace{0.2em} \\
    
    \tagdoc{
        \textbf{Doc 1 (Title: "Indonesia–Timor Leste Commission of Truth and Friendship"):} The Indonesia–Timor Leste Commission on Truth and Friendship was a truth commission established jointly by the governments of \textbf{Indonesia and East Timor} in August 2005...\\
        \textbf{Doc 2 (Title: "Indonesia–Timor Leste Commission of Truth and Friendship"):} ...investigate acts of violence that occurred around the independence referendum held in East Timor in 1999...
    }
    \vspace{0.6em} \\

    \tagthink{From the documents, it is clear that Indonesia has a Commission of Truth and Friendship with East Timor. Now I need to find out if East Timor is a newly declared independent country and who its president is.}
    \vspace{0.4em} \\
    
    \tagsearch{East Timor president}
    \vspace{0.2em} \\
    
    \tagdoc{
        \textbf{Doc 1 (Title: "Francisco Guterres"):} Francisco Guterres, popularly known as Lú-Olo... is an East Timorese politician who has been \textbf{President of East Timor since 2017}...\\
        \textbf{Doc 3 (Title: "José Ramos-Horta"):} ...José Manuel Ramos-Horta... was the President of East Timor from 20 May 2007 to 20 May 2012...
    }
    \vspace{0.6em} \\

    \tagthink{From the documents, it is clear that Francisco Guterres is the current president of East Timor. However, I need to confirm if East Timor is a newly declared independent country.}
    \vspace{0.4em} \\
    
    \tagsearch{East Timor independence}
    \vspace{0.2em} \\
    
    \tagdoc{
        \textbf{Doc 3 (Title: "East Timor"):} ...East Timor became the first new sovereign state of the 21st century on 20 May 2002...
    }
    \vspace{0.6em} \\

    \tagthink{From the documents, it is clear that East Timor gained independence in 2002, which makes it a newly declared independent country. Therefore, the president of the newly declared independent country that is part of the Commission of Truth and Friendship with Indonesia is Francisco Guterres.}
    \vspace{0.4em} \\
    
    \tagans{Francisco Guterres} \\
    
    \bottomrule
    \end{tabularx}
\end{table*}

\end{document}